\documentclass[12pt,conference]{IEEEtran}
\usepackage{tikz}
\usepackage{placeins}  
\usepackage{amsmath}
\usepackage{graphicx}
\usepackage{subcaption}
\usepackage{amsfonts}
\usepackage{color, colortbl}
\usepackage{booktabs}
\usetikzlibrary{positioning, shapes, arrows.meta}
\IEEEoverridecommandlockouts
\usepackage{cite}
\graphicspath{ {plots/} }
\usetikzlibrary{positioning, fit}

\usepackage{forest}
\usepackage{amsmath,amssymb,amsfonts}
\usepackage{algorithmic}
\usepackage{graphicx}
\usepackage{textcomp}
\usepackage{xcolor}
\usepackage{tikz}
\usepackage{multirow}
\usepackage{hyperref}
\def\BibTeX{{\rm B\kern-.05em{\sc i\kern-.025em b}\kern-.08em
		T\kern-.1667em\lower.7ex\hbox{E}\kern-.125emX}}
\makeatletter
\def\ps@IEEEtitlepagestyle{%
	\def\@oddfoot{\mycopyrightnotice}%
	\def\@evenfoot{}%
}
\def\mycopyrightnotice{%
	{\footnotesize manuscript is submitted to \hfill}
	\gdef\mycopyrightnotice{}
}
\begin{document}
	
	\title{Conditional Shift-Robust Conformal Prediction for Graph Neural Network\\
		{
		}
	}
	
	\author{\IEEEauthorblockN{S. Akansha}
		\IEEEauthorblockA{\textit{Department of Mathematics}\\
			\textit{Manipal Institute of Technology}\\
			Manipal Academy of Higher Education - 576104, India.\\
			akansha.agrawal@manipal.edu.}
	}	
	\maketitle	
	\begin{abstract}
This article presents the first systematic study of the impact of distributional shift on uncertainty in Graph Neural Network (GNN) predictions. We introduce Conditional Shift Robust (CondSR), a novel, model-agnostic approach for improving GNN performance under conditional shift\footnote{Representing the change in conditional probability distribution \(P(label|input)\) from source domain to target domain.}. Our two-step method first trains the GNN using a regularized loss function designed to minimize conditional shift in latent stages, refining model predictions. We then apply conformal prediction to quantify uncertainty in the output of the trained GNN model. Comprehensive evaluations on standard graph benchmark datasets demonstrate that CondSR consistently achieves predefined target marginal coverage, enhances the accuracy of state-of-the-art GNN models by up to 12\% under conditional shift, and reduces prediction set size by up to 48\%. These results highlight CondSR's potential to significantly improve the reliability and efficiency of GNN predictions in real-world applications where distributional shifts are common. As the first to systematically address this challenge, our work opens new avenues for robust uncertainty quantification in graph-structured data. Our code\footnote{https://github.com/Akanshaaga/CondSR.git} implementation is publicly available for further exploration and experimentation.

	\end{abstract}
	\section{Introduction}	
	In recent years, the proliferation of data across various domains has spurred a heightened interest in leveraging graph structures to model intricate relationships \cite{ran_she_kout-15a, les_mca-12a, def_bre_van-16a, gil_sch_ril-17a, ham_yin_les-17a}. Graphs, comprising nodes and edges that denote entities and their interconnections, have become a fundamental data representation in domains like social networks \cite{les_mca-12a,che_li_bru-17a,min_gao_pen-21a}, recommendation systems \cite{wan_yuy-22a,gao_wan-22a,chu_yao-22a,che_yeh_wan-22a, gao-zhe-li_23a}, drug discovery \cite{bon-bia-sca_21a, han-lak-liu-21a}, fluid simulation \cite{lin-fot-bha_22a, li-far_22a}, biology \cite{yan_li-23a,jin_eis_son-21a}, and beyond. The increasing diversity and complexity of graph-structured data underscore the necessity for advanced tools to analyze and interpret these intricate relationships, leading to the development of various Graph Neural Networks (GNNs) \cite{kip_wel-2016a, vel_pet_gui-2017a, ham_yin_les-17a, zhu-21a, gas_etal-19a}, which have exhibited remarkable efficacy across a broad array of downstream tasks.

The widespread adoption of Graph Neural Networks (GNNs) across various applications, especially in high-stakes domains such as risk assessment and healthcare, underscores the critical importance of accurately quantifying uncertainty in their predictions. This task is particularly challenging for several reasons. Firstly, traditional GNNs are typically designed to function in in-distribution settings, a condition that is often not met when dealing with real-world data. Secondly, the inherent interconnectedness of entities in graph structures limits the applicability of conventional machine learning methods for assessing prediction confidence. Furthermore, the presence of distributional shifts exacerbates this challenge, introducing additional layers of uncertainty into GNN predictions. As noted by Wan et al. \cite{wan-liu-wan_24a}, distributional shifts significantly contribute to prediction uncertainty, making it imprudent to rely solely on raw GNN outputs, particularly in high-risk and safety-critical applications. Therefore, it is imperative to develop robust methods to assess the confidence in GNN predictions, especially in the presence of distributional shifts.
	
	
Uncertainty quantification in predictive models often involves constructing prediction or confidence sets that encompass a plausible range for the true outcome. While numerous methods have been proposed \cite{hsu-she-tom_22a, zha-kai-han_20a, lak-pri-blu_17a}, many lack rigorous guarantees regarding their validity, particularly concerning the probability that the prediction set covers the actual outcome \cite{ang-bat-mal_20a}. Conformal prediction (CP) has emerged as a prominent tool in this context \cite{sha-vov_08a, ang-bat_21a}. CP leverages historical data to assess the confidence level of new predictions. Formally, for a given error probability \( \epsilon \) and employing a prediction method to generate an estimate \( \hat{y} \) for a label \( y \), CP constructs a set of labels, usually encompassing \( \hat{y} \), with the assurance that \( y \) is included with a probability of \( 1 - \epsilon \). This approach can be employed with any method used to derive \( \hat{y} \). Due to its robust framework, CP has found applications across various domains, including medicine \cite{lei-can_21a, kom-sno-bea_21a, dol-sri-kar_22a, chu-kim-cho_23a}, risk control \cite{bat-ang-lei_21a, ang-bat-fis_22a}, language modeling \cite{qua-fis-sch_23a, deu-alb-mar_24a}, and time series analysis \cite{gib-can_21a, zaf-fer-gou_22a}. However, the application of CP in graph representation learning remains relatively unexplored, primarily due to the inherent interdependencies among entities in graph-structured data. Adapting CP techniques to effectively handle these interdependencies in graph-based models represents a significant and promising area for further research, potentially enhancing the reliability and interpretability of Graph Neural Networks in complex, real-world applications.

\textbf{Present work. }  Conformal Prediction (CP) operates independently of data distribution, relying on the principle of exchangeability, which considers every permutation of instances (in our context, nodes) equally likely. This characteristic positions CP as a robust approach for uncertainty quantification in graph-based scenarios, as it relaxes the stringent assumption of independent and identically distributed (i.i.d.) data. Consequently, CP is particularly well-suited for quantifying uncertainty in prediction models for graph-structured data. In this article, we focus not only on computing prediction sets with valid coverage but also on optimizing the size of these sets. For instance, in a dataset with 7 different classes, a prediction set containing 5, 6, or 7 classes may not be informative, even with 99\% coverage. Ideally, we aim for set sizes smaller than 2, ensuring that the set contains only one class with a predefined relaxed probability of \(1-\epsilon\). This approach enhances the practical utility of our uncertainty estimates. Furthermore, Zargarbashi et al. \cite{zar-ant-boj_23a} establish a broader result demonstrating that semi-supervised learning with permutation-equivariant models inherently preserves exchangeability, further supporting the applicability of CP in graph-based scenarios.

	
This study focuses on quantifying uncertainty in Graph Neural Network (GNN) predictions for node classification tasks, particularly in scenarios characterized by distributional shifts between training and testing data, where \( P_{\text{train}}(Y|X) \neq P_{\text{test}}(Y|X) \). We propose a novel method called Conditional Shift Robust Conformal Prediction (CondSRCP), with the following key contributions:
	\begin{enumerate}
\item 
We analyze how distributional shifts-specifically, conditional shifts-affect the confidence of GNN predictions, using Conformal Prediction (CP) to rigorously quantify uncertainty \footnote{CP quantifies uncertainty by adjusting the size of prediction sets based on model confidence. Under distributional shift, increased uncertainty leads to larger sets, indicating reduced confidence.} and ensure reliability\footnote{CP ensures reliability by maintaining a predefined coverage probability, guaranteeing that the true label is included within the prediction set with high probability, under distributional shifts.} under shift conditions.
\item We introduce a loss function based on our observations of how distributional shifts affect the uncertainty in GNN predictions. This function incorporates a metric designed to minimize distributional shifts during training, resulting in more confident and accurate predictions. When combined with Conformal Prediction, this leads to tighter and more informative uncertainty estimates.
\item Our approach not only enhances the model's prediction accuracy but also achieves improved test-time coverage (up to a predefined relaxation probability \( \epsilon )\) while optimizing the generation of compact prediction sets.
\item CondSR is model-agnostic, facilitating seamless integration with any predictive model across various domains to quantify uncertainty under conditional shift. It preserves and enhances prediction accuracy while simultaneously producing more precise uncertainty estimates. These estimates, in the form of prediction sets, are narrower and more informative across diverse datasets, indicating increased confidence in the model's predictions. Moreover, this method is applicable to unseen graph data, demonstrating its generalizability.
\item Extensive experimentation across various benchmark graph datasets underscores the effectiveness of our approach. 
	\end{enumerate}

	\section{Background and problem framework} \label{sec:background}	
	\noindent\textbf{Notations}
	This work focuses on node classification within a graph in inductive settings. We define an undirected graph \( G = (V, E, X, A) \), consisting of \( |V| = n \) nodes and edges \( E \subseteq V \times V \). Each node \( v_i \in V \) is associated with a feature vector \( x_i \in \mathbb{R}^d \), where \( X \in \mathbb{R}^{n\times d} \) represents the input feature matrix, and the adjacency matrix \( A \in \{0,1\}^{n\times n} \) is such that \( A_{ij} = 1 \) if \( (v_i,v_j) \in E \), and \( A_{ij} = 0 \) otherwise. Additionally, we have labels \( \{ y_i \}_{v_i \in V} \), where \( y_i \in Y = \{1, \ldots, K\} \) represents the ground-truth label for node \( v_i \). The dataset is denoted as \( D := (X,Y) \), initially split into training/calibration/test sets, denoted as \( D_{\text{train}} \), \( D_{\text{calib}} \), and \( D_{\text{test}} \), respectively. In our setting, we generate a training set \( \tilde{D}_{\text{train}} \) from a probability distribution \( P \), while the data \( D \) follows a different probability distribution \( Q \). The model is trained on \( \tilde{D}_{\text{train}} \), with \( D \) serving as the test dataset. This setup allows us to evaluate the model's performance in scenarios where there is a distributional shift between the training and test data, mimicking real-world conditions more closely.
	
	
	\noindent\textbf{Creating Biased Training Data.} Our endeavor involves crafting a training dataset with adjustable bias, a task accomplished through the utilization of personalized PageRank matrices associated with each node, denoted as \( \Pi = (I-(1-\alpha)\tilde{A})^{-1} \). Here, \( \tilde{A} = \mathcal{D}^{-\frac{1}{2}}(A+I)\mathcal{D}^{-\frac{1}{2}} \) represents the normalized adjacency matrix, with \( \mathcal{D} = \sum_{i=1}^{n}a_{ii} \) serving as the degree matrix of \( A+I = (a_{ij}) \) \cite{gas_etal-19a}. The parameter \( \alpha \in [0,1] \) functions as the biasing factor; as \( \alpha \) approaches 0, bias intensifies, and for \( \alpha = 1 \), \( \Pi \) reduces to the normalized adjacency matrix. The crux lies in selecting neighboring nodes for a given target node. Thus, by adhering to the methodology delineated in \cite{zhu-21a}, we harness the personalized PageRank vector to curate training data imbued with a predetermined bias.

	\noindent\textbf{Graph Neural Networks (GNN)} Graph Neural Networks (GNNs) acquire condensed representations that encapsulate both the network structure and node attributes. The process involves a sequence of propagation layers \cite{gil_sch_ril-17a}, where each layer's propagation entails the following steps: (1) \textit{Aggregation}  it iterative aggregate the information from neighboring nodes
	{\small\begin{align}\label{eq:gnnlayer}
			h^{(l)}_u & = UP_{(l-1)}\biggl\{h^{(l-1)}_u, \vspace{1cm} AGG_{(l-1)}\{h^{(l-1)}_v \mbox{where } v \in N_u\}\biggr\}
	\end{align}}
	Here, $h^{(l-1)}_u$ represents the node representation at the $(l-1)$-st layer, typically initialized with the node's feature at the initial layer. $N_u$ denotes the set of neighboring nodes of node $u$. The aggregation function family, denoted as $AGG_{(l-1)}(\cdot)$, is responsible for gathering information from neighboring nodes at the $(l-1)$-st layer and has the form: 
	$$AGG_{(l-1)} : \mathbb{R}^{d_{(l-1)}} \times \mathbb{R}^{d_{(l-1)}} \to \mathbb{R}^{d_{(l-1)}'}$$
	(2) \textit{Update} the update function family, referred to as $UP_{(l-1)}(\cdot)$, integrates the aggregated information into the node's representation at the $(l-1)$-st layer and has the form: 
	$$UP_{(l-1)} : \mathbb{R}^{d_{(l-1)}} \times \mathbb{R}^{d_{(l-1)}'} \to \mathbb{R}^{d_{(l)}}$$
	By iteratively applying this message-passing mechanism, GNNs continuously refine node representations while considering their relationships with neighboring nodes. This iterative refinement process is essential for capturing both structural and semantic information within the graph. For classification task a GNN predicts a probability distribution \( \hat{p}(v) \) over all classes for each node \( v \). Here, \( \hat{p}_{j}(v) \) denotes the estimated probability of \( v \) belonging to class \( j \), where \(j=1,2,\ldots,|Y|\).
	
	\noindent\textbf{Conformal Prediction}	Conformal prediction (CP) \cite{vov-gam-sha_05a} is a framework aimed at providing uncertainty estimates with the sole assumption of data exchangeability. CP constructs prediction sets that ensure containment of the ground-truth with a specified coverage level \(\epsilon\in[0,1]\). Formally, let \( \{(x_i, y_i)\}_{i=1}^m \) be a training dataset drawn from \( X \times Y \), and let \( (x_{m+1}, y_{m+1}) \) be a test point drawn exchangeably from an underlying distribution \( P \). Then, for a predefined coverage level \( 1 - \epsilon \), CP generates a prediction set \( C_{m,\epsilon}(x_{m+1}) \subseteq Y \) for the test input \( x_{m+1} \) satisfying
	\[P\left(y_{m+1} \in C_{m,\epsilon}(x_{m+1})\right) \geq 1 - \epsilon	\]	
	where the probability is taken over \( m + 1 \) data samples \( \{(x_i, y_i)\}_{i=1}^{m+1} \). A prediction set meeting the coverage condition above is considered valid. 
	
Although CP ensures the validity of predictions for any classifier, the size of the prediction set, referred to as \textit{inefficiency}, is significantly influenced by both the underlying classifier and the data generation process (OOD data influence the prediction of the model as well as the coverage and efficiency of CP).
	
Given a predefined miscoverage rate \( \epsilon\), CP operates in three distinct steps: 1. \textit{Non-conformity scores}: Initially, the method obtains a heuristic measure of uncertainty termed as the non-conformity score \( V: X \times Y \rightarrow \mathbb{R} \). Essentially, \( V(x, y) \) gauges the degree of conformity of \( y \) with respect to the prediction at \( x \). 2. \textit{Quantile computation}: The method computes the \( (1-\epsilon) \)-quantile of the non-conformity scores calculated on the calibration set \(D_{\text{calib}}\), that is  calculate \( \tilde{\eta} = \text{quantile}(\{V(x_1, y_1), \ldots, V(x_p, y_p)\}, (1-\epsilon)(1 + \frac{1}{p})) \), where \( p = |D_{\text{calib}}| \). 3. \textit{Prediction set construction}: Given a new test point \( x_{p+1} \), the method constructs a prediction set \( C_{\epsilon}(x_{p+1}) = \{y \in Y : V(x_{p+1}, y) \leq \tilde{\eta}\} \). If \( \{(z_i)\}_{i=1}^{p+1} = \{(x_i, y_i)\}_{i=1}^{p+1} \) are exchangeable, then \( V_{p+1} := V(x_{p+1}, y_{p+1}) \) is exchangeable with \( \{V_i\}_{i=1}^p \). Thus, \( C(x_{p+1}) \) contains the true label with a predefined coverage rate \cite{sha-vov_08a}: \( P\{y_{p+1} \in C(x_{p+1})\} = P\{V_{p+1} \geq \text{Quantile}(\{V_1, \ldots, V_{p+1}\}, 1 - \epsilon) \geq 1 - \epsilon \) due to the exchangeability of \( \{V_i\}_{i=1}^{p+1} \). This framework is applicable to any non-conformity score. We use the following adaptive score in this work.
	
	\noindent\textbf{Adaptive Prediction Set (APS)} utilizes a non-conformity score proposed by \cite{rom-ses-can_20a} specifically designed for classification tasks. This score calculates the cumulative sum of ordered class probabilities until the true class is encountered. Formally, suppose we have an estimator \( \hat{\mu}_j(x) \) for the conditional probability of \( Y \) being class \( j \) at \( X = x \), where \( j = 1, \ldots, |Y| \). We denote the cumulative probability up to the \( k \)-th most promising class as \( V(x, k) = \sum_{j=1}^{k} \hat{\mu}_{\pi(j)}(x) \), where \( \pi \) is a permutation of \( Y \) such that \( \hat{\mu}_{\pi(1)}(x) \geq \hat{\mu}_{\pi(2)}(x) \geq \ldots \geq \hat{\mu}_{\pi(|Y|)}(x) \). Subsequently, the prediction set is constructed as \( C(x) = \{\pi(1), \ldots, \pi(k^*)\} \), where \( k^* = \inf\{ k : \sum_{j=1}^{k} \hat{\mu}_{\pi(j)}(x) \geq \tilde{\eta} \} \).
	
	
	\noindent\textbf{Evaluation metrics }are crucial to ensure both valid marginal coverage and minimize inefficiency. Given the test set \( D_{\text{test}} \), empirical marginal coverage is quantified as Coverage, defined as:
	\begin{equation}
		\text{Coverage} := \frac{1}{|D_{\text{test}}|} \sum_{x_i \in D_{\text{test}}} I(y_i \in C(x_i))
	\end{equation}
	For the classification task, inefficiency is corresponds to the size of the prediction set:
	\begin{equation}
		\text{Ineff} := \frac{1}{|D_{\text{test}}|} \sum_{x_i \in D_{\text{test}}} |C(x_i)|
	\end{equation}
	\textit{A larger interval size indicates higher inefficiency}. It's important to note that the inefficiency of conformal prediction is distinct from the accuracy of the original predictions. 
	
\noindent\textbf{Validity of CP in simultaneous inductive settings}\label{sec:validCP} Conformal Prediction (CP) can be applied to quantify uncertainty in GNN node classification models in both transductive and inductive settings. In transductive settings, Hua et al. \cite{hua-jin-can_24a} demonstrated CP's feasibility based on the exchangeability assumption. They established that node information exchangeability is valid under a permutation invariant condition, ensuring consistent model outputs and non-conformity scores regardless of node ordering. Our study focuses on simultaneous inductive settings, where CP remains applicable under specific conditions. 

\noindent\textbf{Settings for CondSR} In simultaneous inductive settings, the key requirement is that calibration and test data must originate from the same distribution \cite{zar-boj_24a}. Crucially, the non-conformity score in CP remains independent of calibration and test data, preserving exchangeability. Unlike typical GNN training that involves the entire graph, our approach trains the model on a local training set and uses the entire graph as test data. This distinction ensures the applicability of the basic CP model in our simultaneous inductive settings, providing valid uncertainty quantification and coverage.

To illustrate, consider a set of data points $\{(x_i, y_i)\}_{i=1}^p$ and an additional point $(x_{p+1}, y_{p+1})$, all exchangeable. Given a continuous function $V : X \times Y \rightarrow \mathbb{R}$ assessing feature-label agreement and a significance level $\epsilon \in (0, 1)$, we establish prediction sets as $C_{\epsilon}(x_{p+1}) = {y : V(x_{p+1}, y) \leq \tilde{\eta}}$. This ensures that $y_{p+1}$ belongs to $C(x_{p+1})$ with a probability of at least $1 - \epsilon$. Here, $\{(x_i, y_i)\}_{i=1}^p$ are from $\mathcal{D}_{calib}$, and $x_{p+1}$ is from $\mathcal{D}_{test}$, both following the same distribution as the overall data $\mathcal{D}$ but differing from the training data $\tilde{D}_{train}$.
\section{Related Work}\label{sec:relatedwork}

	Several methods in the literature address uncertainty estimation using conformal prediction (CP) under different types of distributional shifts. Tibshirani et al. \cite{tib-foy-can_19a} tackle the problem of covariate shift, where the distribution of input features changes between the source and target domains. Formally, this can be expressed as \(P_{\text{train}}(X, Y) \neq P_{\text{test}}(X, Y)\), assuming that the conditional distribution \(P(Y|X)\) remains consistent during both training and testing phases. They extend CP by introducing a weighted version, termed weighted CP, which enables the computation of distribution-free prediction intervals even when the covariate distributions differ. The key idea is to weight each non-conformity score \(V(x, y)\) by the likelihood ratio \(w(X) = \frac{dP_{\text{test}}(X)}{dP_{\text{train}}(X)}\), where \(P_{\text{train}}(X)\) and \(P_{\text{test}}(X)\) are the covariate distributions in the training and test sets, respectively. The prediction set for a new data point \(x_{n+1}\) is defined as:
	\[ C_{\epsilon}(x_{n+1}) = \{y : V(x_{n+1}, y) \geq \tilde{\eta}\}, \]
	where \(\tilde{\eta}\) is a weighted quantile of the non-conformity scores. This approach ensures that the probability of \(y_{n+1}\) belonging to the prediction set is at least \(1 - \epsilon\).
	
	Gibbs and Candes \cite{gib-can_21a, gib-can_22a} propose adaptive CP for forming prediction sets in an online setting where the underlying data distribution varies over time. Unlike previous CP methods with fixed score and quantile functions, their approach uses a fitted score function \(V_t(\cdot)\) and a corresponding quantile function \(\tilde{\eta}_t(\cdot)\) at each time \(t\). The prediction set at time \(t\) for a new data point \(x_{p+1}^t\) is defined as:
	\[ C_{\epsilon}(x_{p+1}^t) = \{y \in Y : V_t(x_{p+1}, y) \leq \tilde{\eta}_t\}. \]
	They define the realized miscoverage rate \(M_t(\epsilon)\) as:
	\[ M_t(\epsilon) := P\{ V_t(x_{p+1}, y_{p+1}) > \tilde{\eta}_t(1 - \epsilon)\}. \]
	This approach adapts to changing distributions and maintains the desired coverage frequency over extended time intervals by continuously re-estimating the parameter governing the distribution shift. By treating the distribution shift as a learning problem, their method ensures robust coverage in dynamic environments.
	
	Plachy, Makur, and Rubinstein \cite{pla-mak-rub_23a} address label shift in federated learning, where multiple clients train models collectively while maintaining decentralized training data. In their framework, calibration data from each agent follow a distribution \(P^k = P(Y)^k \times P(X|Y)\), with a shared \(P(X|Y)\) across agents but varying \(P^k(Y)\). For \(m\) agents, the calibration data \(\{(x_i^k,y_i^k)\}_{i=1}^{p_k}\) is drawn from a probability distribution \(P^k := P(Y)^k \times P(X|Y)\) for \(k \in \{1,2,\ldots,m\} =: M\), where \(P(X|Y)\), the conditional distribution of the features given the labels, is assumed identical among agents but \(P^k(Y)\), the prior label distribution, may differ across agents. For \(a_k \in \Delta_{|M|}\), where \(\Delta_m\) is the \(|M|\)-dimensional probability simplex, they define the mixture distribution of labels given for \(y \in Y\) as 
	\[ P_y(Y)^{\text{calib}} = \sum_{k=1}^{m} a_k P_y^k(Y). \]
	They propose a federated learning algorithm based on the federated computation of weighted quantiles of agents’ non-conformity scores, where the weights reflect the label shift of each client with respect to the population. The quantiles are obtained by regularizing the pinball loss using Moreau-Yosida inf-convolution.
	
	Comparing these methods highlights their different contexts and applications. Covariate shift methods address static differences between training and test distributions, while adaptive CP focuses on dynamically changing distributions over time. Both methods adjust prediction intervals to maintain coverage but do so in different settings. The label shift approach in federated learning specifically addresses the decentralized nature of federated learning and varying label distributions among clients, differing from both covariate shift and adaptive CP by focusing on label distribution differences and leveraging a federated learning framework.
	
	To the best of our knowledge, there is a lack of methods addressing uncertainty quantification under conditional shift in graph data. This work is pioneering in applying CP to ensure guaranteed coverage in the presence of conditional shift in graph-based data, providing a novel contribution to the field. This advancement underscores the potential for CP to be adapted and extended to a variety of challenging real-world scenarios involving distributional shifts.

	\section{Conditional Shift-Robust GNN}\label{sec:condssr}
	In this section, we propose our Conditional Shift-Robust (CondSR) approach to quantify uncertainty in semi-supervised node classification tasks. This method aims to enhance efficiency while maintaining valid coverage within a user-specified margin, addressing the challenges posed by conditional shifts in graph-structured data.
\subsection{Impact of Distributional Shift on Efficiency}
Conformal Prediction (CP) is designed to maintain a predefined coverage probability, ensuring reliable predictions. However, when a distributional shift occurs—where the test data significantly deviates from the training data—model uncertainty increases. To compensate for this, CP often generates larger prediction sets, incorporating additional potential labels to preserve the desired coverage. While this approach maintains validity, it comes at the cost of reduced efficiency, as the prediction sets become less specific and actionable.

\begin{figure*}[ht]
		\centering
		\begin{tabular}{ccc}
			\multicolumn{1}{c}{Biasing parameter \(\alpha = 0.1\)} &Biasing parameter \(\alpha = 0.4\)&Biasing parameter \(\alpha = 0.6\) \\
			\includegraphics[height=4cm,width=5.6cm]{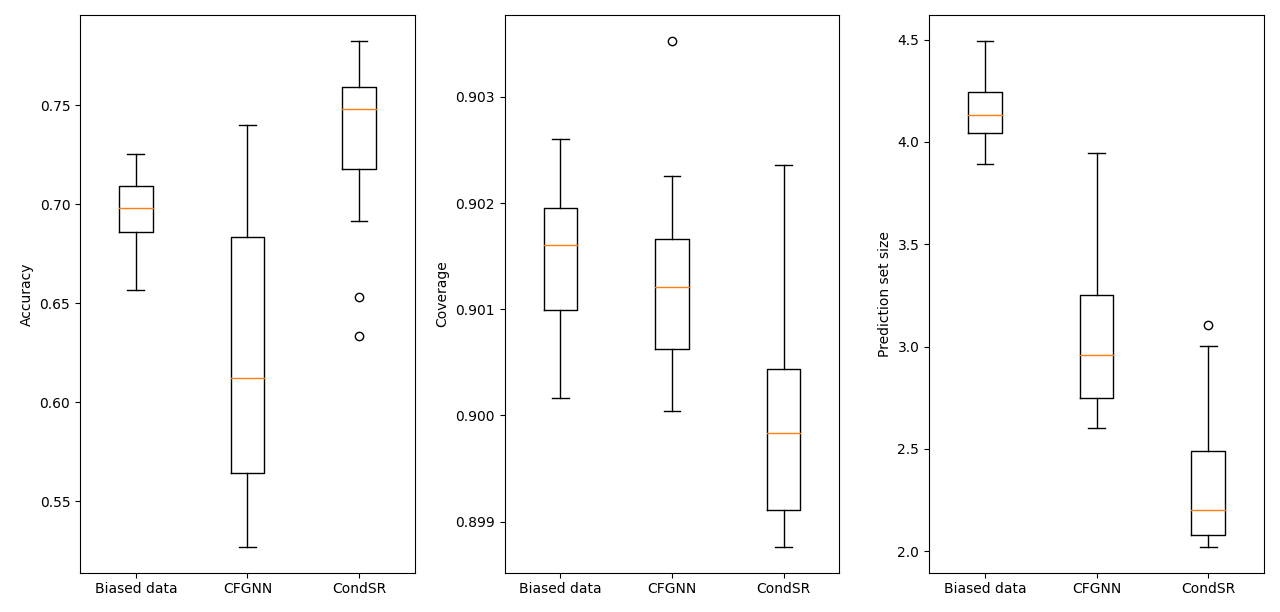} &
			\includegraphics[height=4cm,width=5.6cm]{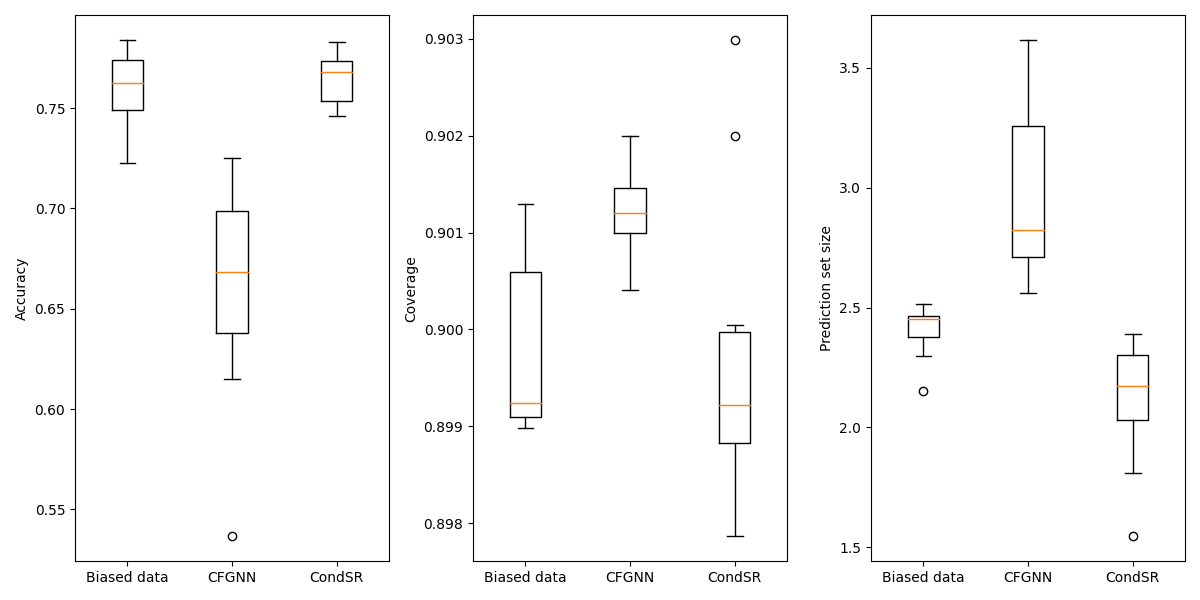} &
			\includegraphics[height=4cm,width=5.6cm]{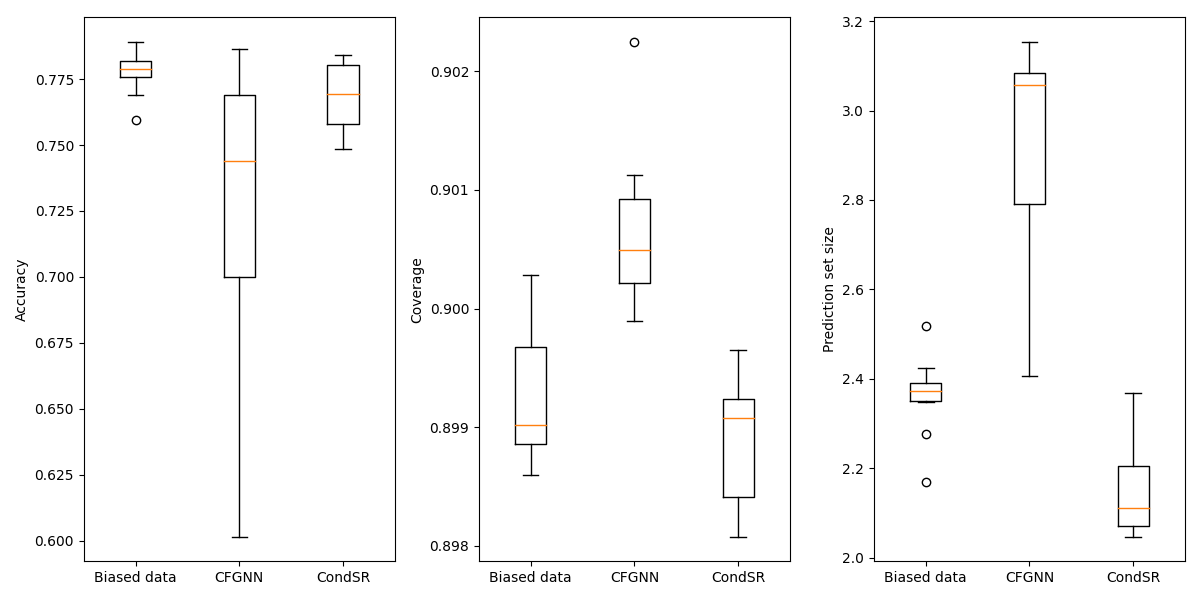}
		\end{tabular}
		\caption{Performance comparison of APPNP, CFGNN, and ConSR under conditional data shift. We evaluate accuracy (left plot in each column), marginal coverage (middle plot in each column), and prediction set size (right plot in each column) across different biasing settings (\(\alpha = 0.1, 0.4, 0.6\)) on the Cora dataset. As \(\alpha\) increases, indicating that the training data distribution is becoming more similar to the test data distribution, both APPNP and ConSR show increased accuracy and decreased prediction set size while maintaining a marginal coverage of 90\%. However, CFGNN does not exhibit a clear pattern in terms of accuracy and prediction set size. ConSR consistently achieves higher accuracy and reduced prediction set size across all scenarios.}	
		\label{fig:cfgnnoncora}
	\end{figure*}

Empirical results in Figure \ref{fig:cfgnnoncora} illustrate that as the intensity of the shift increases, the size of the prediction sets grows significantly, confirming the degradation in efficiency. The underlying reason for this phenomenon lies in how distributional shifts alter the conditional probability distribution, \(P(label | input)\). These shifts disrupt the calibration process of CP, particularly when they lead to increased class overlap or domain mismatch, making it more challenging for the model to assign confident predictions.

A potential strategy to mitigate this efficiency loss is to address distributional shifts during training. By aligning the distributions of the training and test data, the model can reduce uncertainty, resulting in more compact prediction sets. This, in turn, allows for a better balance between accuracy and efficiency.
	\begin{table*}[ht]
		\centering
		\caption{Accuracy comparison of five state-of-the-art (SOTA) GNN models across on citation network benchmarks. The accuracy is measured using $F_1$-micro scores. The transitions are as follows: from IID data to biased data (first arrow) and from biased data to CondSR (second arrow). This table highlights the impact of conditional shift (IID to Biased) and the effect of our proposed CondSR method under conditional shift on accuracy. The relative improvement (\%) of CondSR under conditional shift is indicated on the arrows. The results represent the average and standard deviation of prediction sizes, based on 20 GNN runs. The best accuracy achieved is highlighted in bold. The best and second best improvement is highlighted in \textcolor{blue}{blue} and \textcolor{red}{red}, respectively.}
		\label{table:accuracy}
		\begin{tabular}{l|l|l}
			\hline \\[-0.7em]
			Model & $~~~~~~~~~~~~~~~~~$Cora &$~~~~~~~~~~~~~~~~~$ Citeseer \\\hline \\[-0.7em] \vspace{.1cm}
			&  $~~~~~$IID $\longrightarrow $ Biased $\longrightarrow $ CondSR & $~~~~~$IID $\longrightarrow~ $ Biased $~\longrightarrow $ CondSR \\ \hline \\[-0.7em] \vspace{.1cm}
			APPNP & $84.11\pm 0.7 \xrightarrow{-14.90\%} 71.57\pm 0.4 \xrightarrow{{4.88\%}}$ \textbf{75.06 $\pm$ 2.6 }& $73.81 \pm 0.8 \xrightarrow{-17.04\%} 61.19 \pm 0.8 \xrightarrow{\textcolor{red}{6.66\%}}$ \textbf{65.26 $\pm$ 0.6} \\ \vspace{.1cm}
			DAGNN & $81.41\pm 0.9 \xrightarrow{-11.83\%} 71.78 \pm 0.2 \xrightarrow{0.81\%} 72.36 \pm 0.1$  & $69.62 \pm 0.1 \xrightarrow{-13.58\%} 60.17 \pm 0.9 \xrightarrow{4.94\%} 63.14 \pm 0.5$  \\ \vspace{.1cm}
			GCN & $80.56\pm 0.4 \xrightarrow{-16.76\%} 67.03 \pm 0.6 \xrightarrow{\textcolor{red}{4.96\%}} 70.35\pm 0.5$  & $70.02 \pm 0.4 \xrightarrow{-15.19\%} 59.39 \pm 0.5 \xrightarrow{3.17\%} 61.27 \pm 0.7$  \\ \vspace{.1cm}
			GAT & $81.09\pm 0.2 \xrightarrow{-15.21\%} 68.75 \pm 0.9 \xrightarrow{4.05\%} 71.54 \pm 0.5$  & $69.68 \pm 0.4 \xrightarrow{-12.45\%} 61.02 \pm 4.9 \xrightarrow{3.15\%} 62.92 \pm 0.5$  \\ \vspace{.1cm}
			GraphSAGE & $80.59\pm 0.9 \xrightarrow{-16.67\%} 67.13\pm 0.7 \xrightarrow{\textcolor{blue}{7.38\%}} 72.09 \pm 0.3$  & $69.19 \pm 0.6 \xrightarrow{-15.36\%} 58.58 \pm 1.6 \xrightarrow{\textcolor{blue}{7.70\%}} 63.09 \pm 1.5$  \\
			\hline
		\end{tabular} \\ \vspace{.2cm}
		
		\begin{tabular}{l|l}
			\hline \\[-0.7em]
			Model & $~~~~~~~~~~~~~~~$Pubmed \\\hline \\[-0.7em] \vspace{.1cm}
			&  $~~~~~$IID $\longrightarrow $ Biased $\longrightarrow $ CondSR \\ \hline \\[-0.7em] \vspace{.1cm}
			APPNP &  $79.51\pm 2.2 \xrightarrow{-18.98\%} 56.42 \pm 4.2 \xrightarrow{9.77\%} \mathbf{70.71 \pm 2.9}$ \\ \vspace{.1cm}
			DAGNN &  $80.09\pm 1.2 \xrightarrow{-22.27\%} 62.25 \pm 0.3 \xrightarrow{\textcolor{blue}{11.22\%}} 69.24 \pm 1.5$ \\ \vspace{.1cm}
			GCN &  $76.09\pm 2.5 \xrightarrow{-20.36\%} 60.61 \pm 3.6 \xrightarrow{\textcolor{red}{10.79\%}} 67.15 \pm 1.5$ \\ \vspace{.1cm}
			GAT &  $77.39\pm 1.9 \xrightarrow{-19.27\%} 62.51 \pm 1.2 \xrightarrow{9.46\%} 68.42 \pm 2.5$ \\
			\hline
		\end{tabular}
	\end{table*}
	\begin{table*}[ht]
		\centering
		\caption{This table presents the empirical inefficiency measured by the size of the prediction set for node classification, where smaller numbers indicate better efficiency. The impact of conditional shift in data is shown (from IID to Biased), with Biased indicating data with conditional shift. The table also shows the relative improvement (\%) of CondSR under conditional shift, indicated on the arrows. The results represent the average and standard deviation of prediction sizes, calculated from 8 GNN runs, each with 200 calibration/test splits. The parameter $\epsilon$ is set to 0.1 (Top table) and 0.05 (Bottom table). The best improvement is indicated by blue text and the second best by red.}
		\label{table:efficiency}
		\begin{tabular}{l|l|l}
			\hline \\[-0.7em]
			Model & $~~~~~~~~~~~~~~~~~$Cora &$~~~~~~~~~~~~~~~~~$ Citeseer \\\hline \\[-0.7em] \vspace{.1cm}
			&  $~~~~~$IID $\longrightarrow $ Biased $\longrightarrow $ CondSR & $~~~~~$IID $\longrightarrow~ $ Biased $~\longrightarrow $ CondSR \\ \hline \\[-0.7em] \vspace{.1cm}
			APPNP & $3.6\pm 1.6 \xrightarrow{{8.33\%}} 3.9 \pm 1.4 \xrightarrow{\textcolor{blue}{\textbf{$-46.15\%$}}} \mathbf{2.1\pm 2.6}$ & $2.8 \pm 0.8 \xrightarrow{{9.29\%}} 3.06 \pm 0.8 \xrightarrow{\textcolor{red}{\textbf{$-15.03\%$}}} \mathbf{2.6 \pm 1.6}$ \\ \vspace{.1cm}
			DAGNN & $4.4\pm 0.3 \xrightarrow{{6.82\%}} 4.7 \pm 0.2 \xrightarrow{\textcolor{red}{\textbf{$-42.55\%$}}} 2.7 \pm 1.1$  & $3.6 \pm 0.4 \xrightarrow{{2.78\%}} 3.7 \pm 0.8 \xrightarrow{\textcolor{blue}{\textbf{$-16.22\%$}}} 3.1 \pm 2.5$  \\ \vspace{.1cm}
			GCN & $3.4\pm 1.1 \xrightarrow{14.71\%} 3.9 \pm 1.1 \xrightarrow{-17.95\%} 3.2\pm 1.2$  & $2.9 \pm 0.2 \xrightarrow{17.24\%} 3.4 \pm 0.7 \xrightarrow{-5.88\%} 3.2 \pm 0.4$  \\ \vspace{.1cm}
			GAT & $3.0\pm 0.9 \xrightarrow{22.52\%} 3.7 \pm 1.9 \xrightarrow{-37.84\%} 2.3 \pm 8.5$  & $2.7 \pm 0.6 \xrightarrow{14.81\%} 3.1 \pm 1.3 \xrightarrow{-12.90\%} 2.7 \pm 3.1$  \\ \vspace{.1cm}
			GraphSAGE & $2.8\pm 1.9 \xrightarrow{21.43\%} 3.4 \pm 1.4 \xrightarrow{-41.18\%} 2.00 \pm 1.1$  & $2.8 \pm 1.1 \xrightarrow{10.71\%} 3.1 \pm 1.6 \xrightarrow{-6.45\%} 2.9 \pm 1.3$ \\
			\hline \\[-0.7em] \vspace{.1cm}
			APPNP & $4.4 \pm 1.6 \xrightarrow{{9.09\%}} 4.8 \pm 1.4 \xrightarrow{\textcolor{blue}{\textbf{$-47.92\%$}}} \mathbf{2.5 \pm 5.3}$  & $3.6 \pm 0.9 \xrightarrow{5.56\%} 3.8 \pm 0.8 \xrightarrow{-23.68\%} \mathbf{2.9 \pm 1.8}$  \\ \vspace{.1cm}
			DAGNN & $5.0 \pm 0.3 \xrightarrow{4.33\%} 5.3 \pm 0.3 \xrightarrow{-33.96\%} 3.5 \pm 1.8$  & $4.2 \pm 0.6 \xrightarrow{4.76\%} 4.4 \pm 0.8 \xrightarrow{\textcolor{blue}{\textbf{$-27.27\%$}}} 3.2 \pm 3.2$  \\ \vspace{.1cm}
			GCN & $4.2 \pm 1.4 \xrightarrow{14.29\%} 4.8 \pm 1.3 \xrightarrow{-12.50\%} 4.2 \pm 1.1$  & $3.6 \pm 0.3 \xrightarrow{19.44\%} 4.3 \pm 0.7 \xrightarrow{-11.63\%} 3.8 \pm 0.5$  \\ \vspace{.1cm}
			GAT & $3.7 \pm 1.2 \xrightarrow{24.32\%} 4.6 \pm 2.2 \xrightarrow{-26.09\%} 3.4 \pm 1.5$  & $3.5 \pm 0.9 \xrightarrow{16.29\%} 4.0 \pm 1.3 \xrightarrow{\textcolor{red}{\textbf{$-23.84\%$}}} 3.1 \pm 3.1$  \\ \vspace{.1cm}
			GraphSAGE & $3.5 \pm 1.9 \xrightarrow{25.71\%} 4.4 \pm 2.3 \xrightarrow{\textcolor{red}{\textbf{$-34.09\%$}}} 2.9 \pm 2.1$  & $3.8 \pm 1.7 \xrightarrow{7.89\%} 4.1 \pm 1.9 \xrightarrow{-21.95\%} 3.2 \pm 1.3$  \\
			\hline
		\end{tabular}
	\end{table*}
	\begin{table*}[ht]
		\centering
		\caption{Empirical marginal coverage of node classification for $\epsilon = 0.1$(upper table) and $\epsilon = 0.05$ (lower table). The result takes the average and standard deviation across 8 GNN runs with 200 calib/test splits. The coverage is calculated using CP.}
		\label{table:coverage}
		\begin{tabular}{l|l|l}
			\hline \\[-0.7em]
			Model & $~~~~~~~~~~~~~~~~~$Cora & $~~~~~~~~~~~~~~~~~$ Citeseer \\ \hline \\[-0.7em] \vspace{.1cm}
			&  $~~~~~$IID $\longrightarrow $ Biased $\longrightarrow $ CondSR & $~~~~~$IID $\longrightarrow~ $ Biased $~\longrightarrow $ CondSR \\ \hline \\[-0.7em] \vspace{.1cm}
			APPNP & $90.10 \pm .00 \xrightarrow{+0.05} 90.15 \pm .00 \xrightarrow{-0.08} 90.07 \pm .00$ & $90.10 \pm .00 \xrightarrow{+0.12} 90.22 \pm .00 \xrightarrow{+0.03} 90.25 \pm .00$ \\ \vspace{.1cm}
			DAGNN & $90.13 \pm .00 \xrightarrow{-0.03} 90.10 \pm .00 \xrightarrow{-0.08} 90.02 \pm .00$ & $90.18 \pm .00 \xrightarrow{+0.09} 90.27 \pm .00 \xrightarrow{-0.08} 90.19 \pm .00$ \\  \vspace{.1cm}
			GCN & $90.06 \pm .00 \xrightarrow{+0.13} 90.19 \pm .00 \xrightarrow{+0.03} 90.22 \pm .00$ & $90.07 \pm .00 \xrightarrow{+0.17} 90.24 \pm .00 \xrightarrow{-0.01} 90.23 \pm .00$ \\  \vspace{.1cm}
			GAT & $90.24 \pm .00 \xrightarrow{-0.11} 90.13 \pm .00 \xrightarrow{+0.01} 90.14 \pm .00$ & $90.10 \pm .00 \xrightarrow{+0.05} 90.15 \pm .00 \xrightarrow{+0.02} 90.17 \pm .00$ \\  \vspace{.1cm}
			GraphSAGE & $90.24 \pm .00 \xrightarrow{-0.18} 90.06 \pm .00 \xrightarrow{-0.04} 90.02 \pm .00$ & $90.04 \pm .00 \xrightarrow{+0.08} 90.12 \pm .00 \xrightarrow{+0.08} 90.20 \pm .00$ \\ \hline \\[-0.7em] \vspace{.1cm}
			APPNP & $95.05 \pm .00 \xrightarrow{+0.03} 95.08 \pm .00 \xrightarrow{-0.01} 95.07 \pm .00$ & $95.05 \pm .00 \xrightarrow{+0.12} 95.17 \pm .00 \xrightarrow{+0.04} 95.21 \pm .00$ \\  \vspace{.1cm}
			DAGNN & $95.05 \pm .00 \xrightarrow{+0.09} 95.14 \pm .00 \xrightarrow{-0.09} 95.05 \pm .00$ & $95.12 \pm .00 \xrightarrow{+0.15} 95.27 \pm .00 \xrightarrow{-0.05} 95.22 \pm .00$ \\  \vspace{.1cm}
			GCN & $95.05 \pm .00 \xrightarrow{+0.10} 95.15 \pm .00 \xrightarrow{-0.01} 95.14 \pm .00$ & $95.04 \pm .00 \xrightarrow{+0.23} 95.27 \pm .00 \xrightarrow{-0.05} 95.22 \pm .00$ \\ \hline \\[-0.7em] \vspace{.1cm}
			GAT & $95.20 \pm .00 \xrightarrow{-0.05} 95.15 \pm .00 \xrightarrow{-0.11} 95.04 \pm .00$ & $95.07 \pm .00 \xrightarrow{+0.10} 95.17 \pm .00 \xrightarrow{+0.01} 95.18 \pm .00$ \\  \vspace{.1cm}
			GraphSAGE & $95.19 \pm .00 \xrightarrow{-0.14} 95.05 \pm .00 \xrightarrow{+0.03} 95.08 \pm .00$ & $95.02 \pm .00 \xrightarrow{+0.11} 95.13 \pm .00 \xrightarrow{-0.01} 95.12 \pm .00$ \\ \hline
		\end{tabular}
	\end{table*}
	\begin{figure*}[ht]
		\centering
		\begin{tabular}{cc}
			\multicolumn{2}{c}{Cora} \\
			\includegraphics[width=0.5\textwidth]{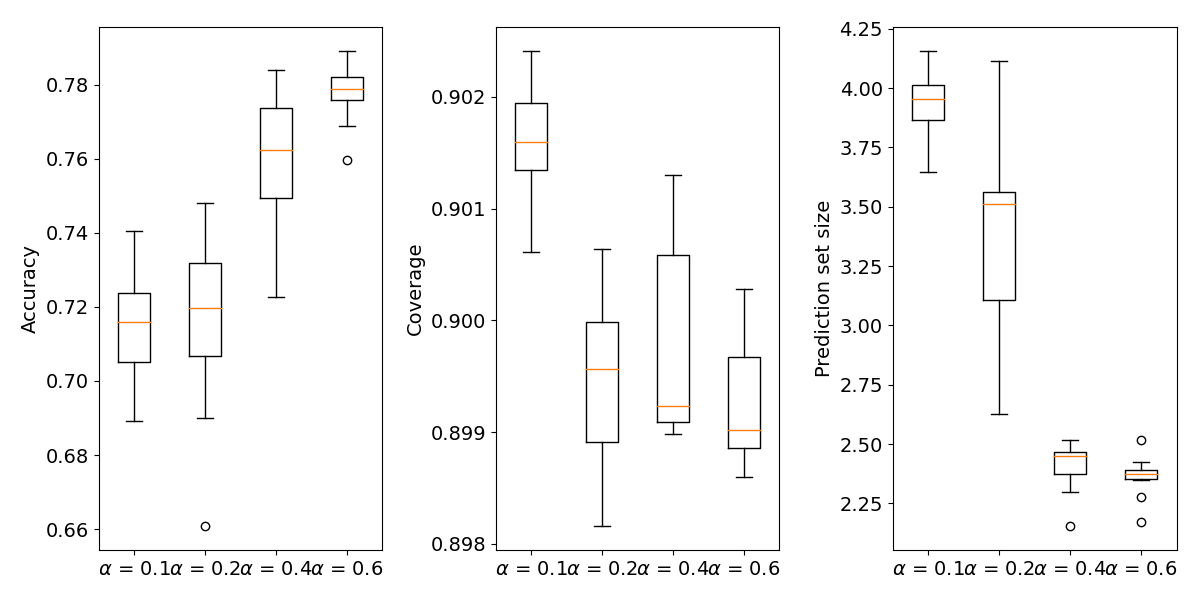} & 
			\includegraphics[width=0.5\textwidth]{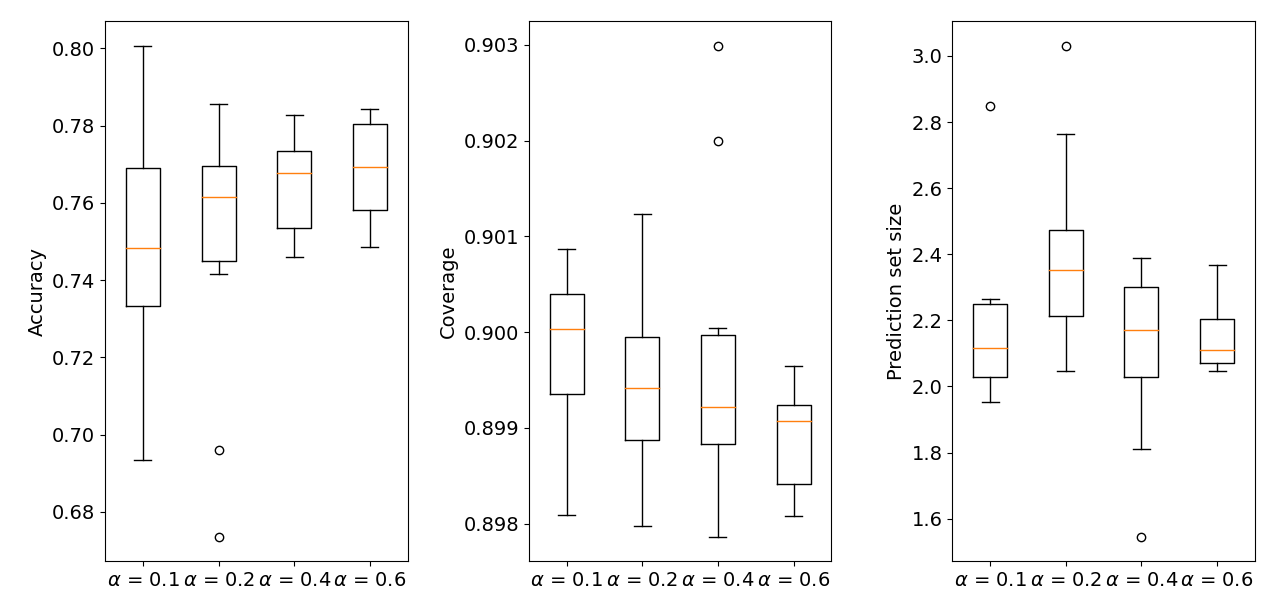} \\
			\multicolumn{2}{c}{Citeseer} \\
			\includegraphics[width=0.5\textwidth]{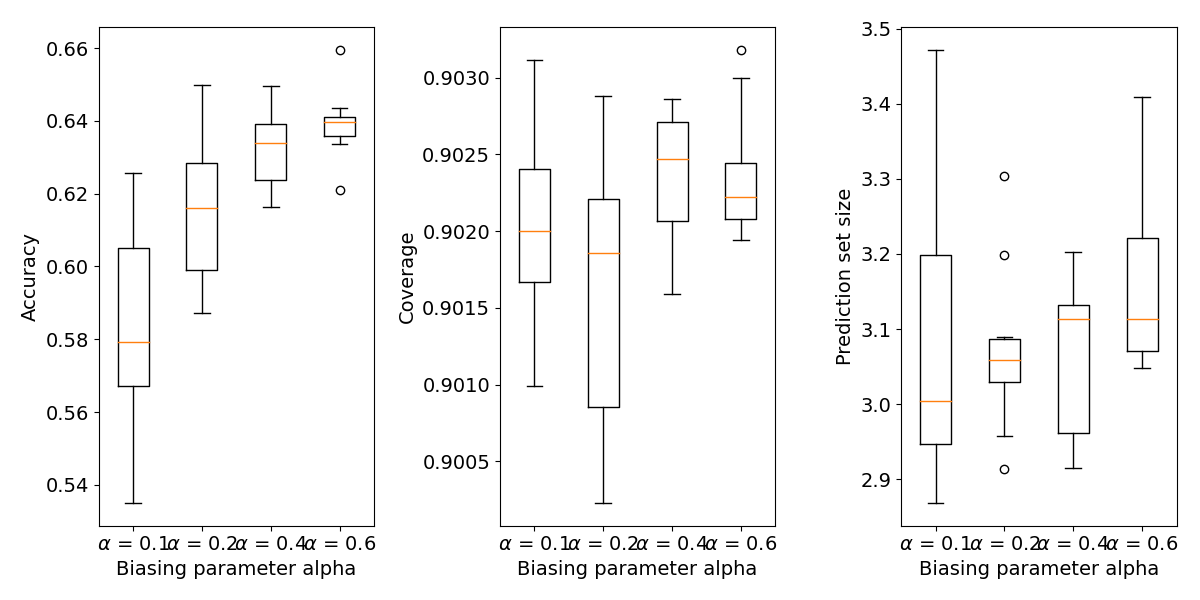} & 
			\includegraphics[width=0.5\textwidth]{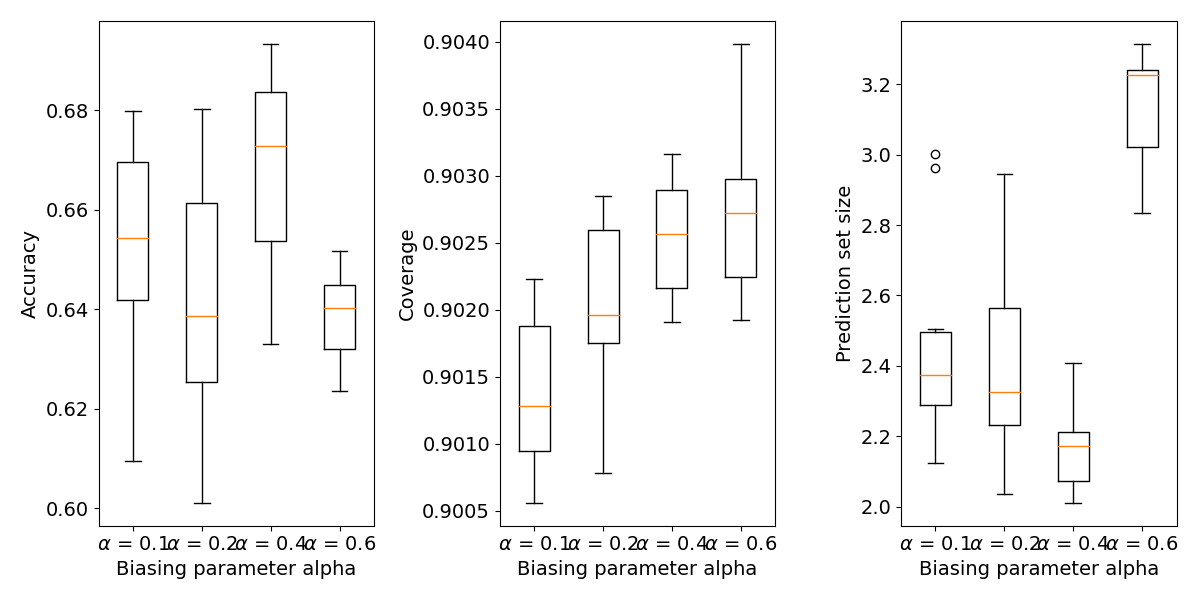}\\
			\multicolumn{2}{c}{Pubmed} \\
			\includegraphics[width=0.5\textwidth]{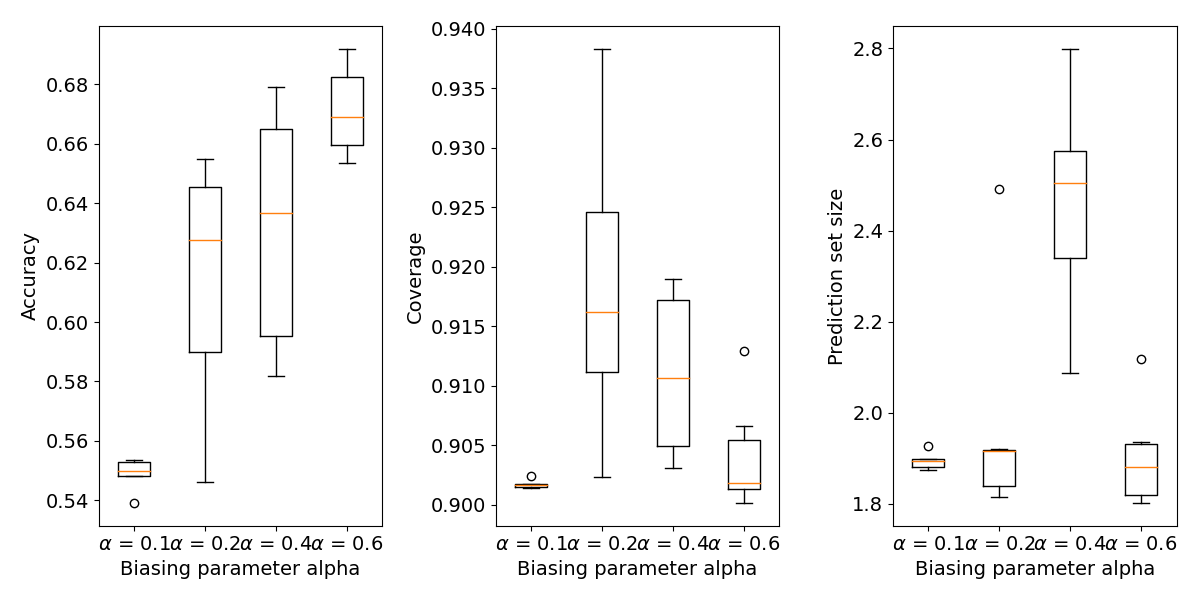} & 
			\includegraphics[width=0.5\textwidth]{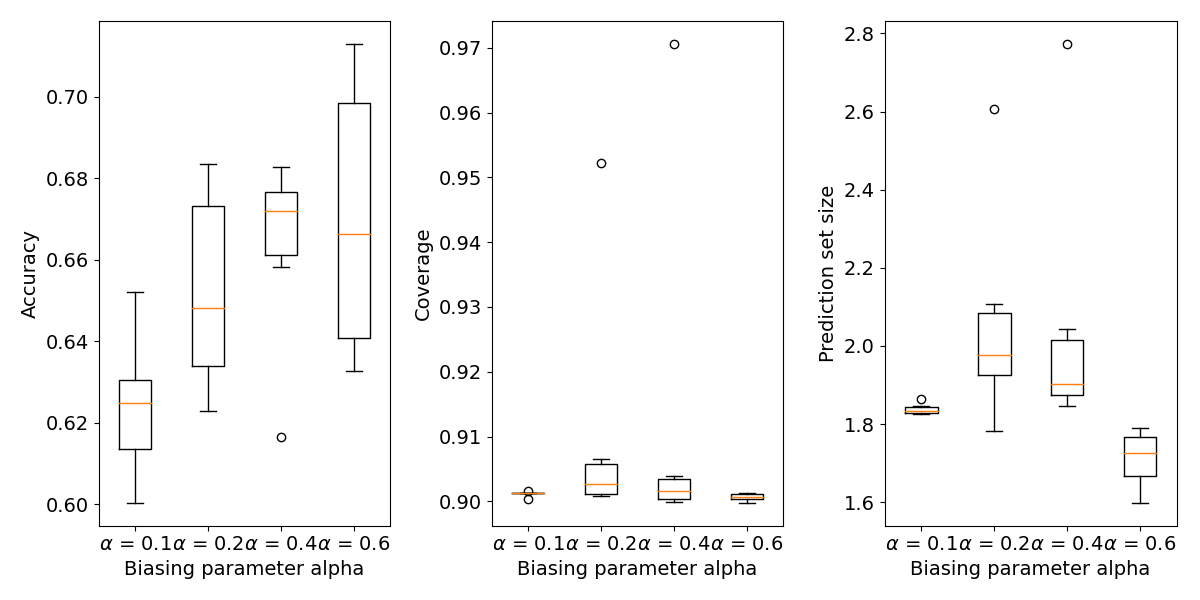} \\
		\end{tabular}
		\caption{The first three plots in each row show the impact of the biasing parameter \(\alpha\) on accuracy (left plot), coverage (middle plot), and size of the prediction set (right plot) under conditional shift in the data. The last three plots in each row show the impact of CondSR on accuracy (left plot), coverage (middle plot), and size of the prediction set (right plot) under the same values of the biasing parameter \(\alpha\).}
		\label{fig:condsr}
	\end{figure*}
\begin{figure*}[ht]
	\centering
	\begin{tabular}{c}
		\multicolumn{1}{c}{Impact of Penalty Parameters \(\lambda_{\text{MMD}}\) and \(\lambda_{\text{CMD}}\) on Accuracy for Cora and Citeseer Datasets} \\
		\includegraphics[height=5.5cm, width=0.2\textwidth]{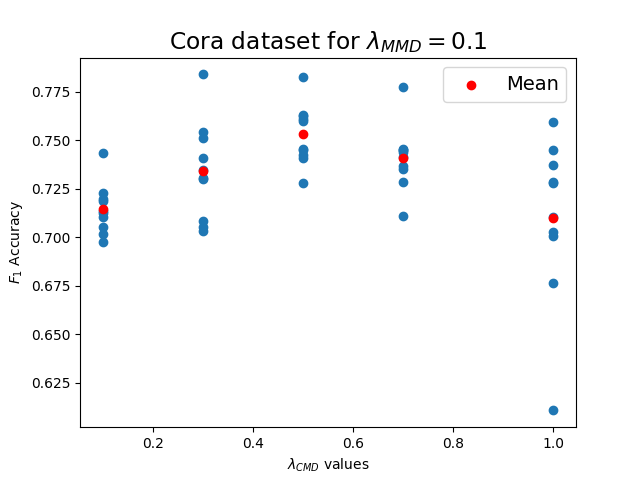}  
		\includegraphics[height=5.5cm, width=0.2\textwidth]{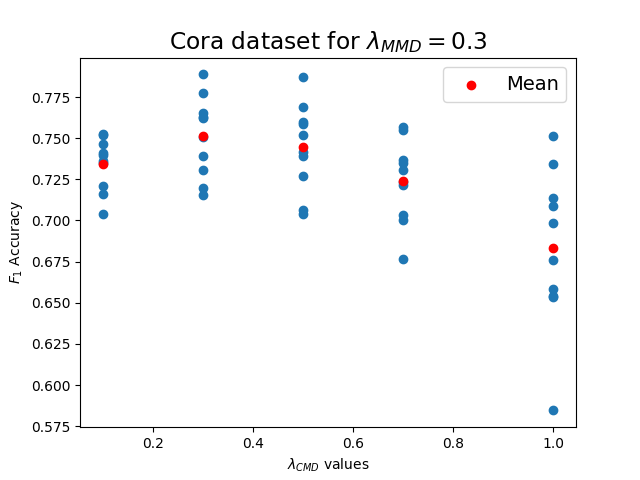} 
		\includegraphics[height=5.5cm, width=0.2\textwidth]{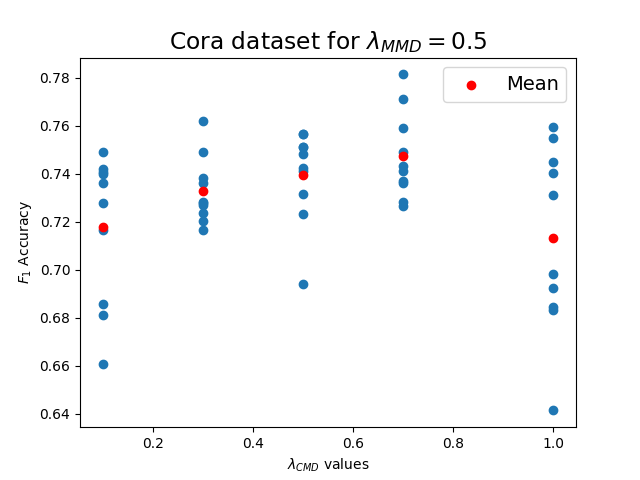}
		\includegraphics[height=5.5cm, width=0.2\textwidth]{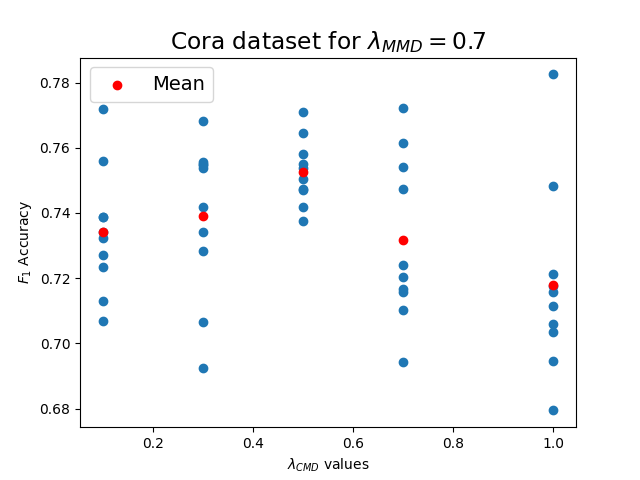}
		\includegraphics[height=5.5cm, width=0.2\textwidth]{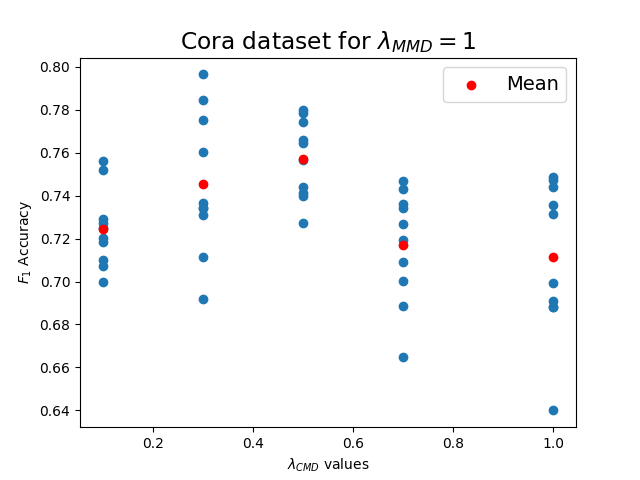}\\
		\includegraphics[height=5.5cm, width=0.2\textwidth]{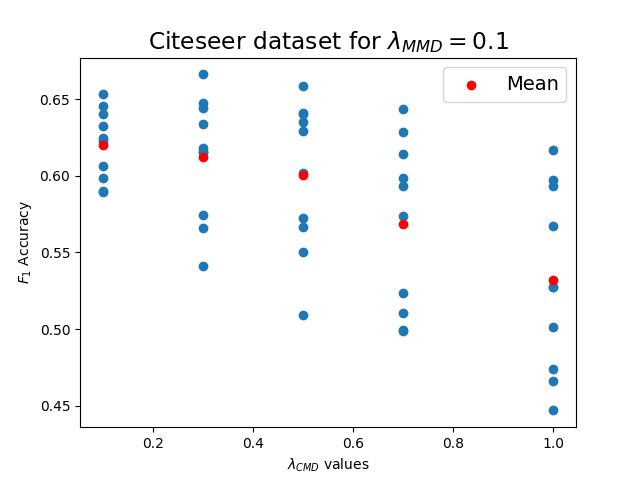} 
		\includegraphics[height=5.5cm, width=0.2\textwidth]{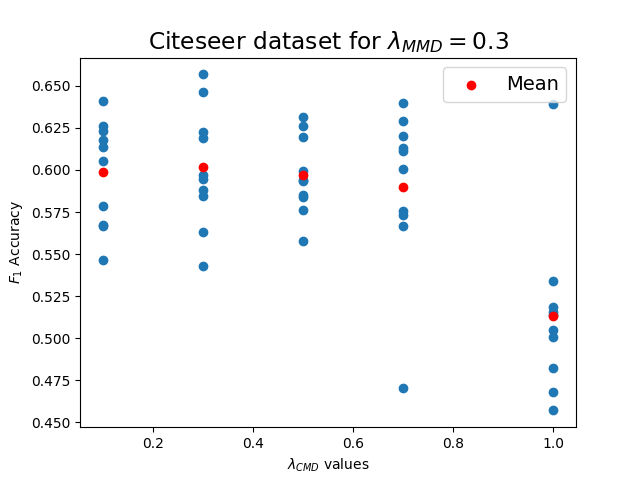} 
		\includegraphics[height=5.5cm, width=0.2\textwidth]{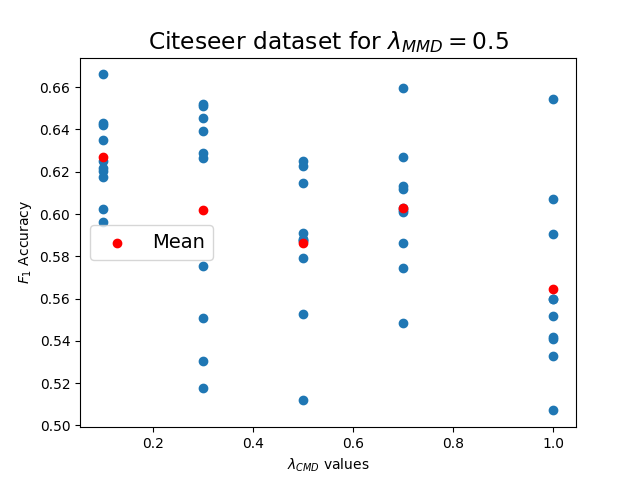}
		\includegraphics[height=5.5cm, width=0.2\textwidth]{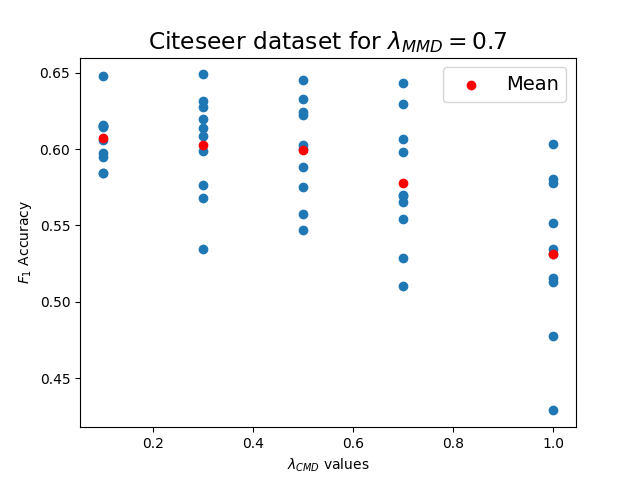}
		\includegraphics[height=5.5cm, width=0.2\textwidth]{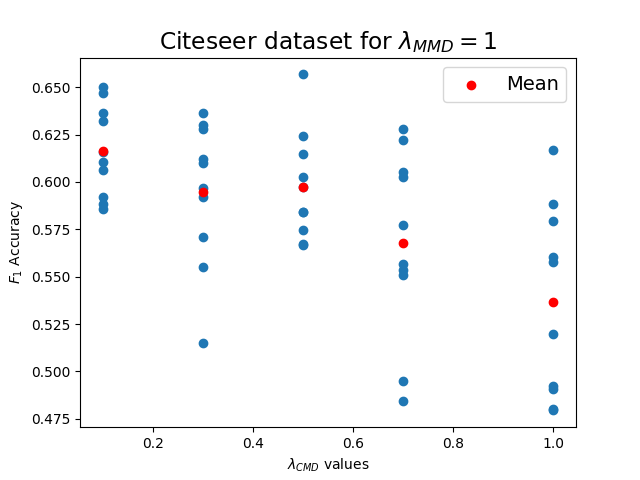}\\
	\end{tabular}
	\caption{The plots in the first and second rows illustrate the effect of varying \(\lambda_{\text{MMD}} = 0.1, 0.3, 0.5, 0.7, 1\) on accuracy for the Cora and Citeseer datasets, respectively. 
		For each fixed value of \(\lambda_{\text{MMD}}\), \(\lambda_{\text{CMD}}\) is varied as \(0.1, 0.3, 0.5, 0.7, 1\). 
		The maximum accuracy is observed at \(\lambda_{\text{CMD}} = 0.5\) and \(\lambda_{\text{MMD}} = 1\) for the Cora dataset, and at \(\lambda_{\text{CMD}} = 0.1\) and \(\lambda_{\text{MMD}} = 0.5\) for the Citeseer dataset.}
	\label{fig:abelation}
\end{figure*}
	

	\subsection{Conditional Shift-Robust Approach}
Our objective is to minimize the conditional distributional shifts between the training and test data distributions. Formally, let \( H = \{h_1, h_2, \ldots, h_n\} \) denote the node representations from the last hidden layer of a GNN on a graph \( G \) with \( n \) nodes. Given labeled data \( \{(x_i, y_i)\} \) of size \( m \), the labeled node representations \( H_l = \{h_1, \ldots, h_m\} \) are a subset of the nodes that are labeled, \( H_l \subset H \). Assume \( H \) and \( H_l \) are drawn from two probability distributions \( P \) and \( Q \). The conditional shift in GNNs is then measured via a distance metric \( d(H, H_l) \).

	Consider a traditional GNN mode \(\Theta\) with learnable function \(f\) with parameter \(\phi\) and \(A\) as adjacency matrix 
	\begin{equation}
		\Theta = f(\phi,A,H),
	\end{equation}
	where \(H :=H_L = f(\phi,A,H_{L-1})\) and \(H_0 = X\). We have \(H_l \in [a,b]^n\) because we use bounded activation function. Let us denote the training samples from \(\tilde{D}_{train}\) as $\{x_i\}_{i=1}^m$, with the corresponding node representations $H_{\text{train}} = \{h_i\}_{i=1}^m$. For the test samples, we sample an unbiased iid sample from the unlabeled data ${X}_{\text{IID}} = \{x'_i\}_{i=1}^m$ and denote the output representations as $H_{\text{IID}} = \{h'_i\}_{i=1}^m$.
	
	To mitigate the distributional shift between training and testing, we propose a regularizer $d : [a, b]^n \times [a, b]^n \rightarrow \mathbb{R}^+$, which is added to the cross-entropy loss, a function quantifies the discrepancy between the predicted labels $\hat{y}_i$ generated by a graph neural network for each node $i$ and the true labels $y_i$. The cumulative loss $l$ is computed as the mean of individual losses across all training instances \(m\). Since $\Theta$ is fully differentiable, we can use a distributional shift metric as a regularization term to directly minimize the discrepancy between the biased training sample and the unbiased IID sample, formulated as follows \cite{zhu-21a,aka_23a}:
	\begin{equation}
		\mathcal{L} = \frac{1}{m} \sum_{i=1}^{m} l(y_i, \hat{y}_i) + \lambda \cdot d(H_{\text{train}}, H_{\text{IID}}),
	\end{equation}
	
Conformal Prediction (CP) provides prediction sets with valid coverage for any black-box prediction model. In the inductive setting, where the model is trained on local training data \(\tilde{D}_{\text{train}}\) and tested on data from \(D_{\text{test}}\), traditional CP can ensure valid coverage but may suffer from inefficiency, as reflected in larger prediction sets. Our CondSR approach enhances the efficiency of CP by aligning the training and testing distributional sifts. During training, by minimizing the distance \(d(H_{\text{train}}, H_{\text{IID}})\), we ensure that the latent representations are robust to conditional shifts. This alignment leads to more compact and reliable prediction sets in the CP framework.	
\subsection{The Framework } Our framework leverages Central Moment Discrepancy (CMD) and Maximum Mean Discrepancy (MMD) to minimize divergence between \( P(y_i|h_i) \) and \( Q(y_i|h_i') \), for \( i=1,2,\ldots,m \), where \( P \) is the probability distribution of the training data \( \tilde{D}_{\text{train}} \) and \( Q \) is the probability distribution of the overall dataset \( D \), which in our setting we consider as IID. To minimize this divergence, we employ CMD and MMD, defined as follows:
\begin{equation}
	\begin{small}
		\begin{aligned}
				d_{CMD}(H_{train},H_{IID}) = & \ \| \mathbb{E}[H_{\text{train}}] - \mathbb{E}[H_{\text{IID}}] \|_2^2 \\
				& + \sum_{k=2}^{K} \| \mathbb{E}[(H_{\text{train}} - \mathbb{E}[H_{\text{train}}])^k] \\
				& - \mathbb{E}[(H_{\text{IID}} - \mathbb{E}[H_{\text{IID}}])^k] \|_2^2,
			\end{aligned}
		\end{small}
	\end{equation}
	\begin{equation}
		d_{MMD}(H_{train},H_{IID}) = \| \mathbb{E}[\phi(H_{\text{train}})] - \mathbb{E}[\phi(H_{\text{IID}})] \|^2,
	\end{equation}
	where \(\phi\) is a feature mapping to a Reproducing Kernel Hilbert Space (RKHS). 
	
	The CMD metric quantifies the disparity between two distributions by comparing their central moments, focusing on aligning higher-order statistical properties. This makes CMD particularly adept at capturing subtle structural variations between distributions that may arise due to non-linear transformations or other complex shifts.	In contrast, the MMD metric operates within a high-dimensional feature space induced by a kernel function, quantifying the dissimilarity between the mean embeddings of two distributions in a rich Hilbert space. MMD is especially effective in detecting broader, mean-level shifts that might not be captured by moment-based comparisons alone. 	By integrating CMD and MMD, we leverage their complementary strengths - CMD addresses higher-order statistical alignment, while MMD ensures robust mean embedding alignment. This combination enhances both accuracy and efficiency compared to relying on either metric in isolation. Empirical results, detailed in Table \ref{table:cmdmmd_vs_condsr}, demonstrate the effectiveness of this dual-metric approach. 
	
	To achieve this, the total loss function combines the task-specific loss (e.g., cross-entropy) with regularization terms based on CMD and MMD:
	\begin{equation}
		\begin{aligned}
		\mathcal{L}_{\text{CondSR}} &= \frac{1}{m} \sum_{i=1}^{m} l(y_i, \hat{y}_i) + \lambda_{\text{CMD}} d_{CMD}(H_{train},H_{IID})\\
		& + \lambda_{\text{MMD}} d_{MMD}(H_{train},H_{IID}),
	\end{aligned} \label{eq:losscondsr}
	\end{equation}
	where \(\lambda_{\text{CMD}}\) and \(\lambda_{\text{MMD}}\) are hyperparameters that balance the trade-off between task performance and distribution alignment. By combining CMD and MMD, the proposed loss function not only mitigates the impact of distributional shifts but also ensures that critical aspects of the distributions are aligned, resulting in improved generalization and predictive performance.
	
	
	\section{Experiments.}\label{sec:expriments} In this section, we validate our CondSR approach to quantify uncertainty in the presence of conditional shift in the data. Specifically, we consider scenarios where the GNN is trained on a local sample from a distribution different from that of the test data (see Section \ref{sec:background} for details). As there is no existing method addressing uncertainty estimation under conditional shift in graph data, we compare the performance of our approach by applying it to baseline models. Our results demonstrate that CondSR provides valid coverage and improves efficiency. Given the crucial role of hyperparameters in the performance of our technique, we conduct a thorough investigation and optimization of these parameters to achieve optimal results.

	\noindent\textbf{Datasets and Evaluation.} Our experimental analysis targets the semi-supervised node classification task using three well-known benchmark datasets: Cora, Citeseer, and Pubmed. To generate biased training samples, we adhere to the data generation methodology outlined in Section \ref{sec:background} and established by \cite{zhu-21a}. For the calibration data, we utilize \(\min \{1000, \frac{|D_{\text{calib}} \cup D_{\text{test}}|}{2}\}\), with the remaining samples designated as test data, assuming that the calibration and test data samples are IID. We perform 200 random splits of the calibration/test sets to estimate the empirical coverage in our experiments.

	
	\noindent\textbf{Baselines.} To evaluate the efficiency of our approach, we compare it with methods that achieve the same coverage, as smaller coverage always results in higher efficiency. Conformal prediction (CP) methods inherently provide exact coverage. To the best of our knowledge, CF-GNN \cite{hua-jin-can_24a} is the CP-based method applicable to any base GNN model, although it is designed for transductive settings. Since no existing method uses CP under conditional shift in graph data, we apply both the CF-GNN model and our CondSR approach under conditional shift to assess performance, depicted in Figure \ref{fig:cfgnnoncora}.  We employ two types of GNN models in our experiments. First, we utilize traditional GNNs that incorporate message passing and transformation operations, including GCN \cite{kip_wel-2016a}, GAT \cite{vel_pet_gui-2017a}, and GraphSage \cite{ham_wil_jur-2017a}. Second, we explore GNN models that treat message passing and nonlinear transformation as separate operations, such as APPNP \cite{gas_etal-19a} and DAGNN \cite{liu_gao_ji-20a}.

	\subsection{Experimental Results} 
	\noindent\textbf{CondSR Enhances Accuracy:} Table \ref{table:accuracy} showcases the F1-accuracy results for semi-supervised node classification. The data clearly indicates a decline in performance for each baseline model under conditional shift (when the model is trained on biased data but tested on IID samples that reflect the real dataset scenario) compared to the in-distribution case (where both training and testing data are IID). Specifically, the average reduction in accuracy is $\approx$15\% on Cora and Citeseer, and 20\% on Pubmed.
	
	In contrast, our proposed technique, CondSR, demonstrates substantial performance improvements under conditional shift. When integrated with baseline models, CondSR consistently boosts their accuracy, as evidenced by the percentage improvements noted on the arrows from biased to CondSR in Table \ref{table:accuracy}. The best average improvement is approximately 8\% on Cora and Citeseer, and approximately 12\% on Pubmed. This consistent enhancement highlights the robustness of our regularization technique across different GNN models under conditional shift conditions.
	
	These findings affirm the effectiveness of CondSR in addressing the challenges posed by distributional shifts, leading to significantly better performance across various GNN models. The results underscore the potential of CondSR to enhance model robustness and accuracy, making it highly valuable for real-world applications where data distribution between training and testing phases may differ.
	
	\noindent\textbf{CondSR Improves Efficiency} Table \ref{table:efficiency} presents the empirical efficiency results for five different GNN models, targeting 90\% and 95\% coverage on two popular citation graph benchmark datasets. The impact of distributional shift on efficiency is demonstrated by the changes from in-distribution settings to distributional shift (indicated by the arrows from IID $\rightarrow$ Biased in Table \ref{table:efficiency}). Under distributional shift, we observe a reduction in efficiency, represented by an increase in the prediction set size, of up to 23\% for 90\% coverage and up to 26\% for 95\% coverage. This significant impact highlights the necessity for model adaptation to maintain confident estimations in the presence of distributional shifts.
	
	Quantifying epistemic uncertainty enables a model to account for its lack of knowledge about unseen data regions, such as when testing data significantly differs from training data. CondSR addresses this challenge effectively. When applied, CondSR consistently improves prediction set size, achieving reductions of up to 48\%. The best average prediction sizes obtained are 2 and 2.5 on Cora, and 2.6 and 2.9 on Citeseer, for nominal levels tuned to achieve 90\% and 95\% empirical coverage, respectively, under conditional shift.
	
	These results underscore the effectiveness of CondSR in enhancing model efficiency under distributional shifts, ensuring more accurate and confident predictions across various GNN models.

	\noindent\textbf{CondSR maintains marginal Coverage:} The coverage outcomes for five distinct GNN models, targeting 90\% and 95\% coverage on two citation graph node classification benchmark datasets, are illustrated in Table \ref{table:coverage}. The findings affirm the preservation of marginal coverage under conditional shift, thus corroborating the theoretical assertions outlined in Section \ref{sec:validCP}. Notably, CondSR consistently upholds marginal coverage across all datasets and baseline GNN models, concurrently augmenting efficiency by minimizing prediction set dimensions and elevating accuracy.

	The empirical results underscore the robustness and effectiveness of the CondSR approach in addressing the challenges posed by distributional shifts. Despite the presence of conditional shifts, CondSR consistently achieves the target coverage levels, validating the theoretical foundations of our approach. Furthermore, CondSR enhances efficiency by reducing prediction set sizes and increasing accuracy, demonstrating substantial improvements over baseline models. The consistent performance improvements across various GNN models and datasets highlight the versatility and adaptability of CondSR. 
	
	\subsection{Ablation}
	In Figure \ref{fig:condsr}, we conduct an ablation analysis to examine the effects of data conditional shift on accuracy, coverage, and prediction set size. We utilize the biasing parameter $\alpha$ to gauge the impact with the APPNP model, comparing the influence of shift with and without the CondSR approach. Initially, we observe that without CondSR, as the shift decreases (indicating a closer alignment of data distribution between training and testing), accuracy increases as anticipated, and prediction set size decreases while maintaining marginal coverage. However, with CondSR, our approach demonstrates improved accuracy and efficiency more robustly, particularly in scenarios with greater distributional shift. As depicted in Figure \ref{fig:condsr}, across all datasets, CondSR exhibits resilience to distributional shift.
	
	\noindent\textbf{Hyperparameters:} The cornerstone parameters within our methodologies are the penalty parameters $\lambda_{CMD}$ and $\lambda_{\text{MMD}}$, corresponding respectively to the losses $\mathcal{L}_{CMD}$ and $\mathcal{L}_{MMD}$. Through meticulous empirical analysis, we have discerned the optimal parameter configurations for the semi-supervised classification task across our benchmark datasets.
	
	For the Cora dataset, peak accuracy is achieved with $\lambda_{CMD} = 0.5$ and $\mathcal{L}_{MMD} = 1$. In the realm of Citeseer, the finest performance emerges with $\mathcal{L}_{CMD} = 0.1$ and $\mathcal{L}_{MMD} = 0.5$, while for the Pubmed dataset, the optimal settings entail $\mathcal{L}_{CMD} = 0.1$ and $\mathcal{L}_{MMD} = 0.1$. Illustrating the comprehensive impact of varying $\lambda_{CMD}$ and \(\lambda_{\text{MMD}}\) values on Cora and Citeseer datasets, Figure \ref{fig:abelation} captures these findings with precision.
	
\noindent\textbf{Effectiveness of Individual Metrics \(d_{CMD}\) and \(d_{MMD}\):}  In the loss function \(\mathcal{L}_{CondSR}\) defined in \eqref{eq:losscondsr}, we incorporate two metrics: \(d_{CMD}\) and \(d_{MMD}\). To justify the use of both metrics, we conducted an empirical study and observed that while both metrics minimize the distance between distributions, each has unique strengths in capturing distinct features of the distributions.  The Central Moment Discrepancy (CMD) metric is particularly effective in aligning higher-order statistical moments, making it well-suited for capturing subtle structural differences between distributions. On the other hand, the Maximum Mean Discrepancy (MMD) metric excels at aligning mean embeddings in a high-dimensional feature space, efficiently detecting broader shifts in distribution. Individually, each metric addresses specific aspects of distributional alignment, but their combination yields superior results in terms of accuracy and efficiency compared to using either metric alone.  
	
Table \ref{table:cmdmmd_vs_condsr} illustrates the impact of the individual metrics and the combined CondSR framework when applied to the APPNP model. We selected the APPNP model for this comparison due to its demonstrated ability to achieve higher accuracy and produce smaller prediction set sizes. This makes it an ideal baseline for evaluating the contribution of \(d_{CMD}\) and \(d_{MMD}\) to the overall performance of CondSR.  

	\begin{table*}[ht]
		\centering
		\caption{This table evaluates the CondSR framework applied to the APPNP model on the Cora and Citeseer datasets. Variants using CMD, MMD and CondSR  are compared in terms of $F_1$-micro accuracy, coverage, and prediction set size. APPNP trained on IID data serves as the baseline, highlighting the complementary strengths and improved performance of the combined metrics within CondSR.}
		\label{table:cmdmmd_vs_condsr}
		\begin{tabular}{l|ccc|ccc}
			\hline \\[-0.7em]
			Model & \multicolumn{3}{c}{Cora} & \multicolumn{3}{|c}{Citeseer} \\\midrule \vspace{.1cm}
			& Accuracy & Coverage & Size & Accuracy & Coverage & Size \\ 
			\midrule
			APPNP (IID) &85.09 $\pm$ $ 0.25$ &$90.10\pm .00$&$3.6\pm1.6$ &75.73 $\pm$ $ 0.30$&$90.10\pm .00$&$2.8\pm 0.8$ \\ \vspace{.1cm}
			APPNP & 70.70 $\pm$ $ 1.9$ &$90.15\pm .00$&$3.9\pm 1.4$& 60.78 $\pm$ $ 1.6$&$90.22\pm .00$&$3.06\pm 0.8$ \\ \vspace{.1cm}
			CondSR-APPNP w. MMD & 71.21 $\pm$ $ 1.0$ &$90.14\pm .00$&$4.1\pm 1.0$& 60.46 $\pm $ $1.9$ &$90.20\pm .00$&$3.1\pm 1.5$\\ \vspace{.1cm}
			CondSR-APPNP w. CMD & 75.03 $\pm$ $ 1.2$ &$89.95\pm .11$&$2.3 \pm 3.7$ & 62.72 $\pm$ $ 1.9$&$90.22\pm .00$& $3.1 \pm 2.3$  \\ \vspace{.1cm}
			CondSR-APPNP (Ours) & $\mathbf{75.14 \pm  1.8}$ &$90.07\pm .00$&$\mathbf{2.1\pm 2.6}$ & $\mathbf{65.01 \pm 2.0}$&$90.25\pm .00$&$\mathbf{2.6\pm 1.6}$  \\
			\hline
		\end{tabular}
	\end{table*}
	\section{Conclusion} In this work, we extend conformal prediction (CP) for GNNs to the inductive setting, where the training and test data distributions differ. Our Conditional Shift-Robust Conformal Prediction framework effectively addresses distributional shifts in GNNs, ensuring robust performance and reliable uncertainty quantification. This approach significantly enhances model robustness and performance under conditional shift conditions, making it a valuable tool for semi-supervised node classification. By maintaining marginal coverage while improving efficiency, our framework ensures more reliable and accurate predictions—an essential capability for real-world applications where data distributions can vary between training and testing phases.
	
	Future research will focus on extending CP to more challenging scenarios, such as extreme distributional shifts and varied application domains. Additionally, integrating CP with techniques to mitigate common GNN limitations, such as over-squashing, over-smoothing, and handling noisy data, will further enhance the robustness and applicability of GNNs in diverse environments.

	\section*{Acknowledgment}
	I extend my heartfelt appreciation to Dr. Karmvir Singh Phogat for providing invaluable insights and essential feedback on the research problem explored in this article. His thoughtful comments significantly enriched the quality and lucidity of this study.
	\section*{Author's Information} Dr. S Akansha, is the only author of this manuscript. 
	
	\noindent\textbf{Affiliation} Dr Akansha. Department of Mathematics, Manipal Institute of Technology, Manipal-576104, India.
	
	\noindent\textbf{Contribution} Dr Akansha: Conceptualization, Methodology, Software, Data curation, Writing- Original draft preparation, Visualization, Investigation, Validation, Writing- Reviewing and Editing.
	
	\section*{Statements and Declarations}
	\textbf{Conflict of Interest} The authors declare that they have no known competing financial interests or personal relationships that could have appeared to influence the work reported in this paper.
	
	\bibliographystyle{./IEEEtran}
	\bibliography{./IEEEabrv,bibfile_UQGNN}
	
\end{document}